\documentclass[10pt,twocolumn,letterpaper]{article}
\usepackage[accsupp]{axessibility} 

\usepackage[applications]{wacv}

\usepackage{graphicx}
\usepackage{amsmath}
\usepackage{amssymb}
\usepackage{booktabs}

\usepackage{xcolor}
\usepackage{multirow}
\usepackage{arydshln}
\usepackage{cancel}
\usepackage{tikz}
\usepackage{rotating}
\usepackage{makecell}
\usepackage{floatrow}
\usepackage{setspace}
\DeclareUnicodeCharacter{2212}{--}

\newfloatcommand{capbtabbox}{table}[][\FBwidth]

\usepackage{xspace}
\makeatletter
\DeclareRobustCommand\onedot{\futurelet\@let@token\@onedot}
\def\@onedot{\ifx\@let@token.\else.\null\fi\xspace}
\def\eg{\emph{e.g}\onedot} 
\def\ie{\emph{i.e}\onedot} 
 
 \def\vs{\emph{vs}\onedot}
\def\wrt{w.r.t\onedot} 

\def\etal{\emph{et al}\onedot}

\usepackage{amssymb}
\usepackage{pifont}
\newcommand{\cmark}{\ding{51}}%
\newcommand{\xmark}{\ding{55}}%

\usepackage[pagebackref,breaklinks,colorlinks]{hyperref}

\usepackage[capitalize]{cleveref}
\crefname{section}{Sec.}{Secs.}
\Crefname{section}{Section}{Sections}
\Crefname{table}{Table}{Tables}
\crefname{table}{Tab.}{Tabs.}

\def\wacvPaperID{554} 
\def\confName{WACV}
\def\confYear{2025}

\begin{document}

\title{Automated Patient Positioning with Learned 3D Hand Gestures}

\author{Zhongpai Gao$^*$, Abhishek Sharma$^*$, Meng Zheng, Benjamin Planche, Terrence Chen, Ziyan Wu \\
United Imaging Intelligence, Boston MA, USA \\
\texttt{\{first.last\}@uii-ai.com} 
}
\maketitle
\def\thefootnote{*}\footnotetext{Equal contribution}

\begin{abstract}
   Positioning patients for scanning and interventional procedures is a critical task that requires high precision and accuracy. The conventional workflow involves manually adjusting the patient support to align the center of the target body part with the laser projector or other guiding devices. This process is not only time-consuming but also prone to inaccuracies. In this work, we propose an automated patient positioning system that utilizes a camera to detect specific hand gestures from technicians, allowing users to indicate the target patient region to the system and initiate automated positioning. Our approach relies on a novel multi-stage pipeline to recognize and interpret the technicians' gestures, translating them into precise motions of medical devices. We evaluate our proposed pipeline during actual MRI scanning procedures, using RGB-Depth cameras to capture the process. Results show that our system achieves accurate and precise patient positioning with minimal technician intervention. Furthermore, we validate our method on HaGRID, a large-scale hand gesture dataset, demonstrating its effectiveness in hand detection and gesture recognition.
\end{abstract}


\section{Introduction}
\label{sec:intro}
\begin{figure*}[t]
    \centering
    \includegraphics[width=1.0\textwidth]{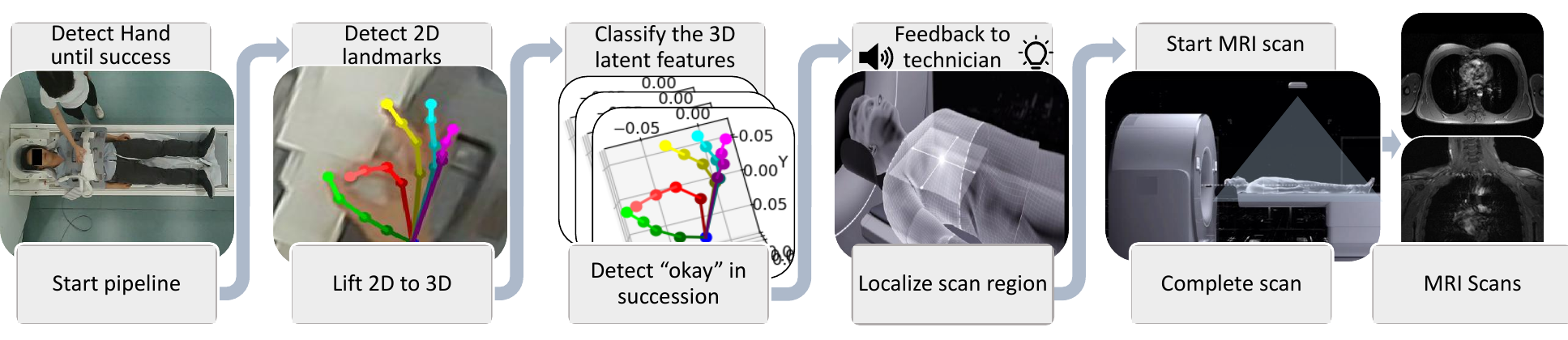}
    \caption{Illustration of the proposed end-to-end automated patient positioning workflow.
    }
    \label{fig:systemworkflow}\vspace{-6pt}
\end{figure*}
Patient positioning is a central aspect of medical care that can greatly impact the success of imaging or surgical procedures, as it impacts both patient safety and the final effectiveness of the procedures. However, this process is traditionally performed through manual measurements and visual estimates, which are not only time-consuming but also lead to errors and inconsistencies. Recent advancements in computer vision have provided opportunities to develop more accurate and efficient patient positioning techniques to streamline the whole process. 

Hand gestures provide intuitive and effective non-verbal communication with computer interfaces \cite{kaur2016review}. Hand gesture recognition can combine with other computer vision techniques, \eg, human body detection and pose estimation, to streamline the process of patient positioning. This can be particularly useful in medical imaging applications, such as magnetic resonance imaging (MRI). Traditionally, technicians manually adjust the position of the patient support to ensure that the center of the scanning region is properly aligned with the laser projector on the MRI scanner. However, by using hand gesture recognition, technicians can simply post the gesture on the desired scan center once they finished placing the coil on the patient. Once the gesture is detected and confirmed by the camera above the patient support, the system can quickly and accurately locate the scanning region and trigger the movement of the bed into the gantry so the center of the target body part aligns with the ISO center. This can significantly improve the workflow and efficiency of patient positioning. This workflow can be straightforwardly adapted to other medical scanners (\eg, CT, X-ray, and angiography) or treatment and surgery procedures where it may be challenging for physicians to manually operate and control medical devices.  

This paper proposes a hand gesture recognition method for patient positioning in scan rooms (SR) that utilizes images from ceiling-mounted cameras. However, several challenges need to be addressed for accurate recognition in such settings. 
First, hands have numerous degrees of freedom, \eg, different hand orientations, and may be subject to occlusions and self-similarity. Second, ceiling-mounted cameras must have a large field of view to monitor the working area. As a result, each hand covers only a tiny region of the image, \ie, typically with a resolution of around $40 \times 40$px. This makes their detection and hand gesture recognition difficult. Third, lighting conditions can vary significantly across different scan rooms, posing a challenge to vision systems, particularly under low-light conditions. At last, the 2D appearance of hands and gestures can greatly vary when observed from different angles, making hand gesture recognition based on hand images or 2D landmarks ambiguous and unreliable. 

To address these challenges, we propose several novel components in the analytics pipeline. First, we use orientation- and association-aware hand detection to center-align the hand with the predicted hand bounding box and rotation angle and associate it with its body (technician or patient). This reduces the degrees of freedom of the target data/distribution and helps with recognition accuracy. Second, we use data augmentations for hand landmark detection to address the issues of small-object detection and low-light conditions in scan rooms. Specifically, we simulate low resolution and adjust brightness and contrast to increase the diversity of the training set. Finally, we use a dual-modality 3D pose estimation model to resolve the ambiguity of hand gestures from 2D space. These techniques help to improve the accuracy and robustness of the gesture recognition system, allowing it to locate the scanning region and trigger the movement of the bed to the desired position efficiently. On a dataset collected in realistic clinical environments, we demonstrate the efficacy and reliability of the proposed workflow. The main contributions of this work can be summarized as follows:

\begin{itemize}
\item We introduce a new automated patient positioning workflow and system based on 3D hand gesture recognition in scan rooms, which can improve overall efficiency and throughput compared to traditional manual workflow, and facilitate contactless medical scans.

\item We propose a multi-stage pipeline for 3D hand gesture recognition, relying on a novel orientation- and association-aware hand detection model, illumination-invariant 2D landmark detection model, and a dual-modality 3D pose and gesture recognition model. Together, they ensure robust and accurate patient-positioning in challenging clinical settings.

\item We conduct extensive experiments on our clinical benchmark collected from real clinical environments and a large-scale public hand gesture dataset. The results demonstrate that the proposed system and method outperforms off-the-shelf methods.  
\end{itemize}


\section{Related Work}

\noindent\textbf{Hand gesture recognition.} Hand gesture recognition can be categorized into static and dynamic use cases \cite{reifinger2007static,pisharady2015recent}. Static gestures focus on hand shapes from single images, while dynamic gestures require a series of hand movements to be recognized. Although dynamic gesture recognition \cite{yuanyuan2021review} is often preferred in applications that require complex and natural interactions, static gesture recognition \cite{oyedotun2017deep, kapitanov2024hagrid} is typically more accurate and easier to implement. Therefore, we choose static gestures to ensure precise and accurate patient positioning control.

Traditionally, data gloves are used to interpret gestures in human-computer interaction (HCI) \cite{dipietro2008survey}. Data gloves can easily provide the locations, orientations, and other configurations of the palm and fingers by using sensors attached to the gloves. However, this approach requires users to wear the glove physically \cite{laviola1999survey}, which is not convenient and adds an extra cost for the data glove. Thus, we adopt a computer vision approach that only uses an RGB camera for our 3D hand gesture recognition. 

The computer-vision based approach provides contactless communication between human and computers \cite{kaur2016review}. Early RGB-based methods \cite{4587752, prisacariu2011robust} use low-level visual cues, \eg, edges, silhouettes, and optical flow, to fit hand models. However, this is extremely challenging due to various lighting conditions, complex backgrounds, heavy occlusions. For example, Perimal \etal \cite{perimal2018hand} provide 14 gestures under controlled-conditions room lighting using an HD camera at short distances (0.15 to 0.20m). The recognition rate drops significantly under low-light or high-noise conditions. 

Deep learning-based approaches are the current state-of-the-art. Some methods \cite{adithya2020deep, tan2021hand} use convolutional neural networks to classify the gestures directly. However, these methods do not generalize well in the wild. 3D reconstruction and pose estimation methods \cite{cai2018weakly, ge20193d, gao2020semi} can regress the 3D coordinates of keypoints, which can further be used for hand gesture recognition. However, the processing time of these methods is the bottleneck for real-time applications.

MediaPipe hand \cite{zhang2020mediapipe} is a real-time on-device hand tracking solution that predicts hand keypoints from a single RGB camera. Many gesture recognition methods \cite{harris2021applying, veluri2022hand, halder2021real} have been proposed based on MediaPipe hand. However, MediaPipe hand is designed for hands close to the camera (around 1m or less) and does not work well for ceiling cameras in scan rooms where hands are usually small. 


\noindent\textbf{AI-based patient positioning.} Recent advancements in computer vision have proven invaluable for patient positioning, including patient detection and pose estimation in operating rooms (ORs) \cite{srivastav2018mvor}, human pose estimation on low-resolution depth images in ORs \cite{srivastav2019human}, physics-guided in-bed pose estimation \cite{liu2019seeing}, and patient positioning based on patient body modeling in scan rooms \cite{karanam2020towards,zheng2022self,Singh2017DARWIN}. These methods show great promise in improving patient positioning. However, it's imperative to recognize the inherent limitations associated with methods reliant on human pose. First, their efficacy is compromised under occlusions, such as when patients are enshrouded with sheets or coils. Second, the identified joints, though accurately detected, might not necessarily correspond to the exact location intended for scanning. In light of these challenges, the proposed 3D hand gesture-based system significantly enhances the accuracy of locating the body part and initiates the scanning process more effectively.

\section{Methodology}
We present the overall pipeline and then detail our orientation- and association-aware hand detection, domain augmentation, and bi-modal 3D pose regression for gesture recognition. 

\begin{figure*}[t!]
    \centering
    \includegraphics[width=1.0\textwidth]{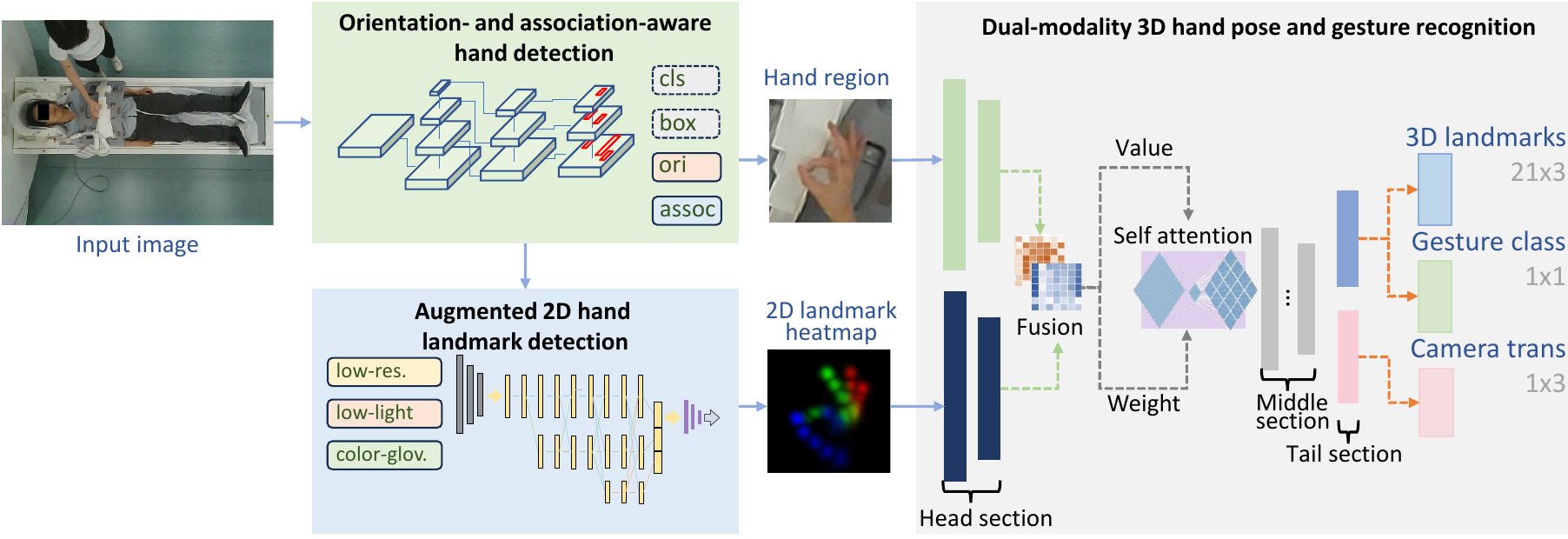}
    \caption{Overall analytics pipeline.
    }
    \label{fig:pipeline}\vspace{-6pt}
\end{figure*}

\subsection{Automated Positioning Workflow and System}
Take MRI as an example: after placing the coil on the patient, traditionally, the technician would visually locate the scanning center on the body part and manually move the patient support until the center of the body part aligns with the laser projector on the gantry. 
We propose to automate this positioning, as illustrated in Figure \ref{fig:systemworkflow}. 
Once the coil is placed, the technician post an \textit{``OK''} hand gesture on top of the coil with the center of the circle formed by the thumb and index (\textit{``O''}) aligned with the center of the target body part. Once the system detects the \textit{``OK''} gesture from the technician, the system will enter a confirmation phase, where audio and lighting prompts are activated to notify the user that the patient support control is pending confirmation. 
If the technician's intention was not to position the patient, they could withdraw the gesture to cancel the workflow. Otherwise, if they maintain the gesture for the entire confirmation phase (\eg, 3 seconds), the system then sends the control signal to start the motion of patient support. In practice, during confirmation, the system shall detect a succession of multiple \textit{``OK''} gestures to activate, making the whole workflow more reliable. After confirmation, the pixel location of gesture \textit{``O''} center is extracted from the image
and projected to the MRI coordinate system as target position, using the corresponding depth value from the depth sensor and camera-system calibration data obtained during setup (\ie, the rigid transform from camera to MRI system frame). The center of the target body part is thus automatically aligned with the ISO-center of the MRI system, before the scanning starts. Note that \textit{``OK''} gesture used in this work is just one concrete example and can be replaced with any other gesture to fulfill various requirements.  s.  

In our system, the analytics pipeline runs on an edge device that continuously processes the images from the ceiling-mounted camera above the patient support. The edge device with limited GPU/CPU memory operates on a low-power mode, restricting the GPU's computational capabilities. The system employs a workflow that wakes up the edge device from sleep to conduct periodic technician detection to optimize efficiency without overheating the edge device.

\subsection{Analytics Pipeline}
Fig.\ \ref{fig:pipeline} illustrates the overall analytics pipeline of our system. First, an orientation- and association-aware hand detection model locates hands in the image frame, producing a bounding box, class, rotation angle, and association vector for each. The rotation angle represents the angle relative to the hand's upward direction and the association vector links the hand center to its body center. Second, each technician's hand region (the patient's hands are filtered out according to the association prediction) is cropped and aligned based on the predicted bounding box and rotation angle. Third, a 2D hand landmark detection model generates the heatmap of 2D landmarks, which is then passed to a 3D hand pose estimation network, along with the cropped hand image, to regress the 3D hand landmarks. Finally, a gesture classification model utilizes the latent feature of the 3D pose estimation model to predict whether the gesture is \textit{``okay''} or \textit{``non-okay''}.

\paragraph{Orientation- and association-aware Hand Detection.}
\label{sec:detection}

For hand alignment, inspired by \cite{sung2021device}, we define the hand orientation based on the 2D landmarks to limit the degrees of freedom. The vector linking the middle base knuckle to the wrist estimates the rotation angle. The upward of this vector is set as canonical direction (\ie, zero angle) to make the hand upward. Thus, hand landmark detection and gesture recognition models do not need to cover various hand orientations. For hand-body association, we define an association vector as the hand center to its body center as in \cite{gao2024pbadet}. We assume the person lying in the bed is the patient (\ie, the bounding box intersection of the body and bed over the body bounding box is larger than 0.65). Therefore, we can filter out the patient's hands according to the association vector and the system responds exclusively to the technician's gestures to prevent unintended interactions.

We propose a novel extended object representation that integrates hand orientation and association to construct a single-stage orientation-and-association-aware hand detector. Our detector directly infers a set of objects as $o = \{o_b, o_c, o_r, o_d\}$, where $o_b = \{l, t, r, b\}$ denotes the coordinates of the left-top and right-bottom bounding-box corners;  $o_c = \{c_1, c_2\}$ is the classification result (\ie, labels of hand and body); $o_r$ is the rotation angle; and $o_d = \{d_x, d_y\}$ relates to the hand-body association. Our simple yet efficient innovation has at least two advantages: 1) the orientation and association learning task can enforce the detection model to learn hand shapes; 2) the output prediction includes the bounding boxes, orientation angles, and hand-body association, which avoids multi-stage predictions or tedious post-processing for hand alignment and association. 

\paragraph{Augmented 2D Landmark Detection.}
\label{sec:landmark}

In scan rooms, hand regions in a monitoring camera have low resolution and are under low-lighting conditions. To train the hand 2D landmark detection model accordingly, we perform data augmentation to simulate these conditions, \ie, for each training image $\mathbf{I}_{ori}$ we generate a low-resolution version 
$$
    \hat{\mathbf{I}}_{lr} = \mathbf{K}^s_{\uparrow}\left(\min\left(\mathbf{K}^s_{\downarrow}\left(1, \mathbf{b}\cdot\mathbf{I}_{ori}\right)\right) + \mathbf{n}\right),
$$
where $\mathbf{K}^s_{\downarrow}$ and $\mathbf{K}^s_{\uparrow}$ represent downsampling and upsampling operations of scale $s \in \{1, 2, 4, 8\}$, $\mathbf{n}$ is Gaussian noise, and $\mathbf{b} \in(0.75, 1.25)$ is a ratio randomly sampled to adjust the image brightness. In addition, technicians may wear different color gloves. To simulate color-gloved hands, we leverage SAM \cite{kirillov2023segment} to segment the hand region using the bounding box as a prompt and overlap random colors in the training set. We train the 2D hand landmark detection model that takes aligned hand images as input and outputs the heatmaps of 2D landmarks.

\paragraph{Dual-modality 3D Pose Estimation for Gesture.}
\label{sec:gesture}
Gesture recognition based on 2D landmarks is inherently ambiguous as hands are highly articulated and look very different from varying perspectives. In contrast, hand gestures are more consistent in 3D space. We propose a gesture recognition framework based on hand 3D pose estimation, as shown in Fig. \ref{fig:pipeline}. We take the aligned hand image and the heatmap from the 2D landmark detection model as inputs. We extend the MobileOne \cite{mobileone2022} to support two inputs and include a Hadamard self-attention fusion module. In detail, we divide the network after the second stage and before the last stage into head, middle, and tail sections. Then, we duplicate the head section for the two inputs and the tail section for multiple outputs.

The two head sections extract feature maps of the hand image and landmark heatmap as $\mathbf{X}_{rgb}\in\mathbb{R}^{C\times H\times W}$ and $\mathbf{X}_{lmk}\in\mathbb{R}^{C\times H\times W}$, respectively. To fuse the two modalities, we add a Hadamard self-attention fusion module. First, we concatenate the features maps as $\mathbf{X}_{rgb|lmk}\in\mathbb{R}^{2C\times H\times W}$. Then, we use two separate convolution blocks to get the value $\mathbf{V}_{rgb|lmk}$ and weight $\mathbf{W}_{rgb|lmk}$. At last, we compute the Hadamard product of the value and weight as
$\hat{\mathbf{X}}_{rgb|lmk} = \mathbf{V}_{rgb|lmk} \odot \mathbf{W}_{rgb|lmk}$ to fuse the feature maps, then passed through the middle section. This can be summarized as:
$$
\hat{\mathbf{X}}_{att} = f_{\theta_{v}}(\mathbf{X}_{cat}) \odot f_{\theta_{w}}(\mathbf{X}_{rgb|lmk}),
$$ 
with $\mathbf{X}_{rgb|lmk} = \mathbf{X}_{rgb} \Vert \mathbf{X}_{lmk}$,
where $f$ are convolution blocks with parameters $\theta_{v}$ and $\theta_{w}$ respectively, $\odot$ and $\Vert$ are resp. the Hadamard and concatenation operators.

The two tail sections output two latent features as $\mathbf{F}_{l}\in\mathbb{R}^{1280}$ responsible for the local 3D landmarks and gesture and $\mathbf{F}_{g}\in\mathbb{R}^{1280}$ responsible for the global camera translation. Following the latent features, three two-layer MLP [1280, 96, $C_{out}$] are used to predict the 3D landmarks ($C_{out}=21\times3$), gesture class ($C_{out}=1$), and camera translation ($C_{out}=1\times3$). We adopt the local latent feature $\mathbf{F}_{l}$ for gesture recognition as gestures are only related to the local 3D landmarks and are independent of the global camera translation.


\section{Experiments}

\subsection{Datasets}
We consider three public datasets, COCO-WholeBody \cite{xu2022zoomnas},  HanCo \cite{ZimmermannAB21}, and HaGRID \cite{kapitanov2024hagrid}. COCO-WholeBody extends COCO \cite{lin2014microsoft} with whole-body annotations, including
face and hand bounding boxes, as well as the keypoints of face, hand, and foot. We train our hand detection and landmark detection models on this dataset. HanCo \cite{ZimmermannAB21} is a structured collection of hand images that were recorded in short video clips with a calibrated and time-synchronized multi-view camera setup, with annotations of hand segmentation, 3D hand pose and shape, and camera calibration. We thus use this dataset to train our 3D pose estimation model.

In addition, we evaluate our method on HaGRID \cite{kapitanov2024hagrid}, a recent large-scale hand gesture dataset. HaGRID contains 554,800 images with bounding box and gesture annotations, collected from 37,583 subjects. The gesture has 18 classes in total. F1-score and mAP are reported as the metrics. 

To train our two-layer MLP gesture classifier on the local latent feature of the 3D pose estimation model, we build a gesture dataset collected from ceiling cameras in scan rooms. 
We manually label the hand images with two classes: \textit{``non-okay''} and \textit{``okay''}, resulting in 44,151 and 5,719, respectively.
To evaluate the final gesture recognition performance, we create a gesture benchmark collected from ceiling cameras in MRI scan rooms consisting of 65 volunteers as patients and 53 technicians with different coils under various illumination conditions. 
We manually label the camera frames with \textit{``non-okay''} and \textit{``okay''} classes. For the \textit{``okay''} images, we also label the pixel coordinates of the center of the \textit{``okay''} gestures to locate the scanning region of the patient. The image numbers for \textit{``non-okay''} and \textit{``okay''} are 2,155 and 2,067, respectively.

\begin{table*}[t]
\centering
\caption{Gesture recognition and localization evaluation. Note that the 3D-kpt classifier for MediaPipe (F and G) takes the 3D landmarks from MediaPipe as input and is trained on our clinical dataset.}
\label{tab:compare}
\resizebox{1\textwidth}{!}{
\begin{tabular}{c|cc|c|ccc|c|cc}
\toprule
&\multirow{2}{*}{align} & \multirow{2}{*}{augm} & \multirow{2}{*}{\thead{classifier\\input}}
&\multicolumn{3}{c|}{\textit{``ok''}} & \multicolumn{1}{c|}{\textit{``non-ok''}} 
& \multirow{2}{*}{AUC} & \multirow{2}{*}{FPS} 
\\
&&&&Det(\%) & TP(\%) & RMSE(px\#) & TN(\%) \\ \midrule
A&\xmark & \xmark & 2D-kpt & 99.8 & 75.22 & 3.9 (1554)  & 95.82 
& .906 & 15.4
\\
B&\cmark & \xmark & 2D-kpt & 99.8 & 77.06 & 3.8 (1592) & 96.24 
& .909 & 15.4
\\
C&\cmark & \cmark & 2D-kpt & 99.8 & 77.21 & 3.9 (1596) & 97.21 
& .914 & 15.4
\\
D&\cmark & \cmark & 3D-kpt & 99.8 & 67.53 & 3.6 (1396) & 98.01 
& .916 & 13.7
\\
E&\cmark & \cmark & \textbf{3D-feat} & 99.8 & 88.43 & 4.1 (1828) & 95.91 
& \textbf{.948} & 14.4
\\
\midrule
F&\multicolumn{2}{c|}{MediaPipe~\cite{zhang2020mediapipe}+classifier} & 3D-kpt & 57.6 & 35.55 & 10.2 (735) & 93.54 
& .754 & \textbf{17.4}
\\
G&\multicolumn{2}{c|}{YOLOv7+MediaPipe~\cite{zhang2020mediapipe}+classifier} &3D-kpt & 99.8 & 45.47 & 14.1 (940) & 95.31 
& .788 & 13.8
\\
\bottomrule
\end{tabular}}
\end{table*}

\begin{figure}[t]
    \centering
    \includegraphics[width=.85\textwidth]{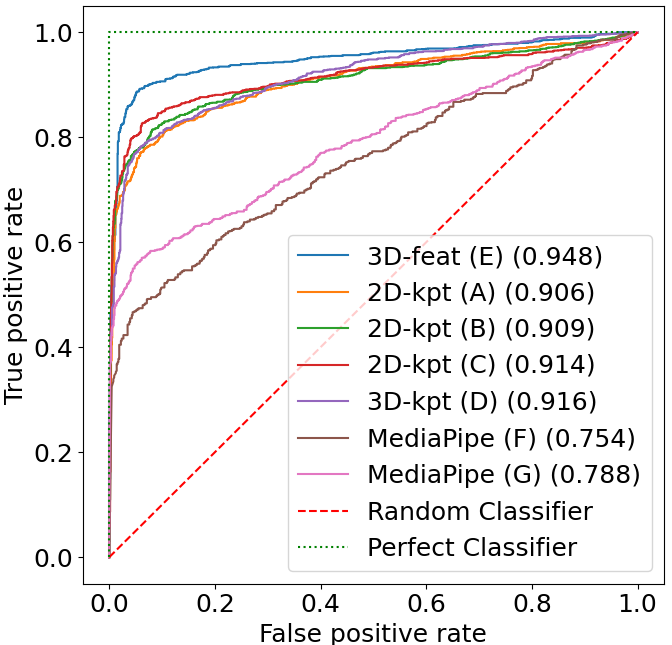}
    \caption{ROC curves of different protocols. }
    \label{fig:roc}
\end{figure}

\subsection{Training}
\noindent{\textbf{Hand detection:}} We choose YOLOv7 \cite{wang2022YOLOv7} as the base model and the COCO-WholeBody dataset \cite{xu2022zoomnas} to train our single-stage anchor-based orientation- and association-aware hand detection model. Since the COCO-WholeBody dataset provides the 2D landmark of hands, we can calculate the orientations from the 2D landmark, as described in Section \ref{sec:detection}.
We train our model with the default settings of \cite{wang2022YOLOv7} but with total epochs of 300, batch size of 64, and input size of $448 \times 448$.

\noindent{\textbf{2D landmark detection:}} We choose HRNetv2 \cite{WangSCJDZLMTWLX19} as the base model and the COCO-WholeBody dataset \cite{xu2022zoomnas} to train our augmented 2D landmark detection model. Since we have the hand orientation from the detection model, we train the 2D landmark detection model with aligned hand images and augment the rotation in the range of [-5, 5] degrees. Additionally, we apply low-resolution and low-light augmentations, as described in Section \ref{sec:landmark}. Except for the alignment and augmentation mentioned above, we train our model with the default settings of \cite{mmpose2020}: total epochs of 210, batch size of 32, and input size of $256 \times 256$. 

\noindent{\textbf{3D pose estimation based gesture recognition:}} We extend the MobileOne \cite{mobileone2022} as the backbone to train our 3D pose estimation based gesture recognition model, as described in Section \ref{sec:gesture}. The whole model is trained in two stages. First, we train the 3D pose estimation model on the HanCo dataset \cite{ZimmermannAB21} using Adam optimizer with a learning rate of $1e^{-4}$, the total epoch 100, batch size 60, and input size of $224\times 224$. In the second stage, we freeze the weight of the 3D pose estimation model and train the two-layer MLP of gesture recognition on our custom gesture dataset using the same settings but a batch size of 512. 

\subsection{Evaluation of the Proposed Method}

\begin{figure*}[th]
    \centering
    \includegraphics[width=1.0\textwidth]{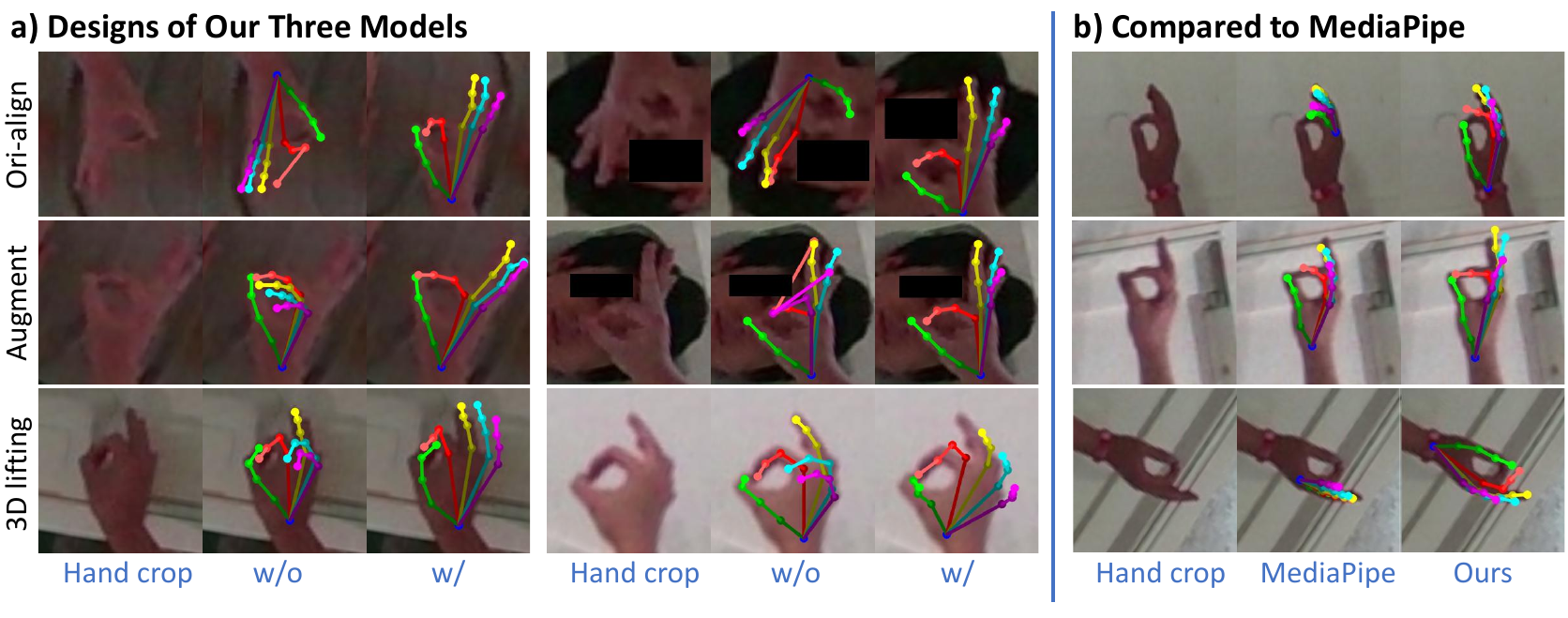}
    \caption{Qualitative results of: a) w/ and w/o three designs: orientation- and association-aware hand detection, augmented landmark detection, and dual-modality 3D hand pose and b) comparison of MediaPipe and our method.}
    \label{fig:quality}
    \vspace{-1em}
\end{figure*}

\begin{figure*}[th]
    \centering
    \includegraphics[width=1.0\textwidth]{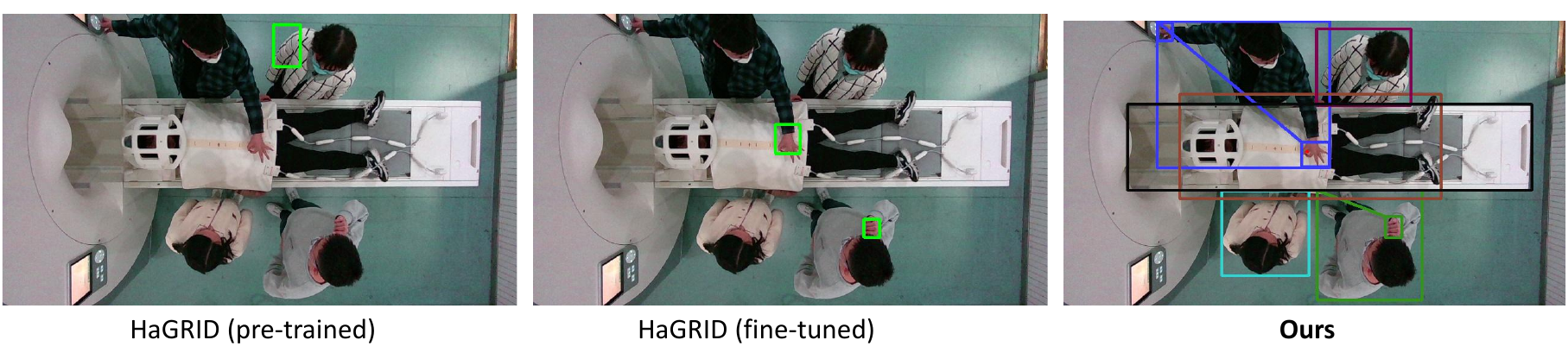}
    \caption{Qualitative comparison with HaGRID \cite{kapitanov2024hagrid} on our gesture benchmark. Note, our hand detection model can associate hands with the body. We use the same color bounding boxes and connect the top-left corners with the same color line to denote hand-body association.}
    \label{fig:hagrid}
    \vspace{-0.6em}
\end{figure*}

\subsubsection{Ablation Study}

In our experiments, we compare our methods over numerous protocols, as shown in Table \ref{tab:compare} and Figure \ref{fig:roc}. We consider five metrics: hand detection rate (Det),  true positive (TP), RMSE of localization errors in pixels \textbf{only} for the detected \textit{``okay''} hands, true negative (TN), area under the ROC curve (AUC), and frames per second (FPS). Note, the threshold is set as $T=0.5$ for TP and TN; Det and AUC are calculated on the whole image level, \ie, any hand detected for Det and the maximum of the predictions for AUC when multiple hands are in an image. Figure \ref{fig:quality} shows the qualitative results to demonstrate the effectiveness of our novel proposals under various lighting conditions.

\noindent{\textbf{Hand orientation alignment.}} From rows A-B of Table \ref{tab:compare}, we can see that the proposed orientation alignment improves the performance on true positive, true negative, and AUC. Since we use the same hand-detection model, the proposed orientation alignment can improve hand landmark detection and gesture classification. Orientation alignment increases the average localization error increases slightly because it detects more \textit{``okay''} hands. The first row of Figure \ref{fig:quality}\textcolor{red}{a} shows some qualitative examples where orientation alignment improves hand landmark detection. 

\noindent{\textbf{Low-resolution and low-light augmentations.}} As presented in rows B-C of Table \ref{tab:compare}, the proposed data augmentation improves the performance in all five metrics, which means our augmentation techniques are useful in our task. The second row of Figure \ref{fig:quality}\textcolor{red}{a} shows some qualitative examples of augmentations improving hand landmark detection.

\noindent{\textbf{Local latent feature of 3D pose estimation.}} From rows C-E of Table \ref{tab:compare}, we compare different gesture classifiers that are trained on different hand data: 2D keypoints (2D-kpt), 3D keypoints (3D-kpt), and latent feature of 3D hand pose model (3D-feat). The true positive of 3D-kpt decreases compared to 2D-kpt while AUC is slightly better. Since the 3D keypoints estimation partially relies on 2D keypoints, the errors are amplified in 3D space when some 2D keypoints are inaccurate. Instead of using 3D keypoints, we use the local latent features of 3D hand pose model, which improve the true positive by \textbf{11.22\%} compared to 2D-kpt. The average localization error increases because more \textit{``okay''} hands are detected (1,828 \vs 1,596), especially more challenging ones. The third row of Figure \ref{fig:quality}\textcolor{red}{a} shows some qualitative examples where the projected 3D keypoints show better results, which means the dual-modality 3D hand pose estimation model is able to improve the input 2D keypoints. 

\begin{table}[htbp]
  \centering
  \caption{Methods of generating the 2D landmark heatmap to train the 3D hand pose model on HanCo dataset. We compare results with heatmaps generated from the 2D landmark ground truth, and those predicted by the 2D landmark detection model, respectively. }
    \begin{tabular}{l|cc|c|c}
    \toprule
       Heatmap   & \multicolumn{2}{c|}{\textit{``okay"}} & \multicolumn{1}{c|}{\textit{``non-okay"}} & \multicolumn{1}{c}{\multirow{2}[2]{*}{AUC}} \\
          & \multicolumn{1}{c}{TP(\%)} & \multicolumn{1}{c|}{RMSE(px\#)} & \multicolumn{1}{c|}{TN(\%)} &  \\
    \midrule
    GT & 77.02 & 5.9 (1592) & 93.97   & .883 \\
    Detector &  \textbf{88.43}  &  \textbf{4.1} (1828)  &   \textbf{95.91}   & \textbf{.948} \\
    \bottomrule
    \end{tabular}%
  \label{tab:heatmap}%
\end{table}%
\noindent{\textbf{Landmark heatmap for 3D pose estimation.}} Our 3D pose estimation model takes two inputs: the RGB hand image and the 2D landmark heatmap. To generate the 2D landmark heatmaps for the HanCo dataset, we use the predicted landmarks from our 2D landmark detection model rather than the ground truth 2D landmarks. Table \ref{tab:heatmap} shows the comparison results, which highlight the benefits of using the predicted 2D landmark heatmap \wrt all four metrics. This approach has three benefits: 1) Using ground truth 2D landmark heatmaps would heavily bias the 3D hand pose model towards the ground truth 2D landmarks, making the fusion of the two modalities inefficient; 2) The 2D landmark heatmap generated by the detection model includes information on the confidence level of each keypoint based on its visibility and visual quality on the input image. This confidence information helps the model balance its reliance on the RGB image and the heatmap. For example, when the confidence level of a keypoint is low, the model will rely more on the input image to correct the 2D landmark predictions from the 2D landmark detection model; 3) During the inference stage, we do not have the ground truth 2D landmarks. Thus, using the predicted 2D landmark heatmap at the training stage is consistent with the inference stage. 

\begin{figure*}[t]
    \centering
    \includegraphics[width=0.83\textwidth]{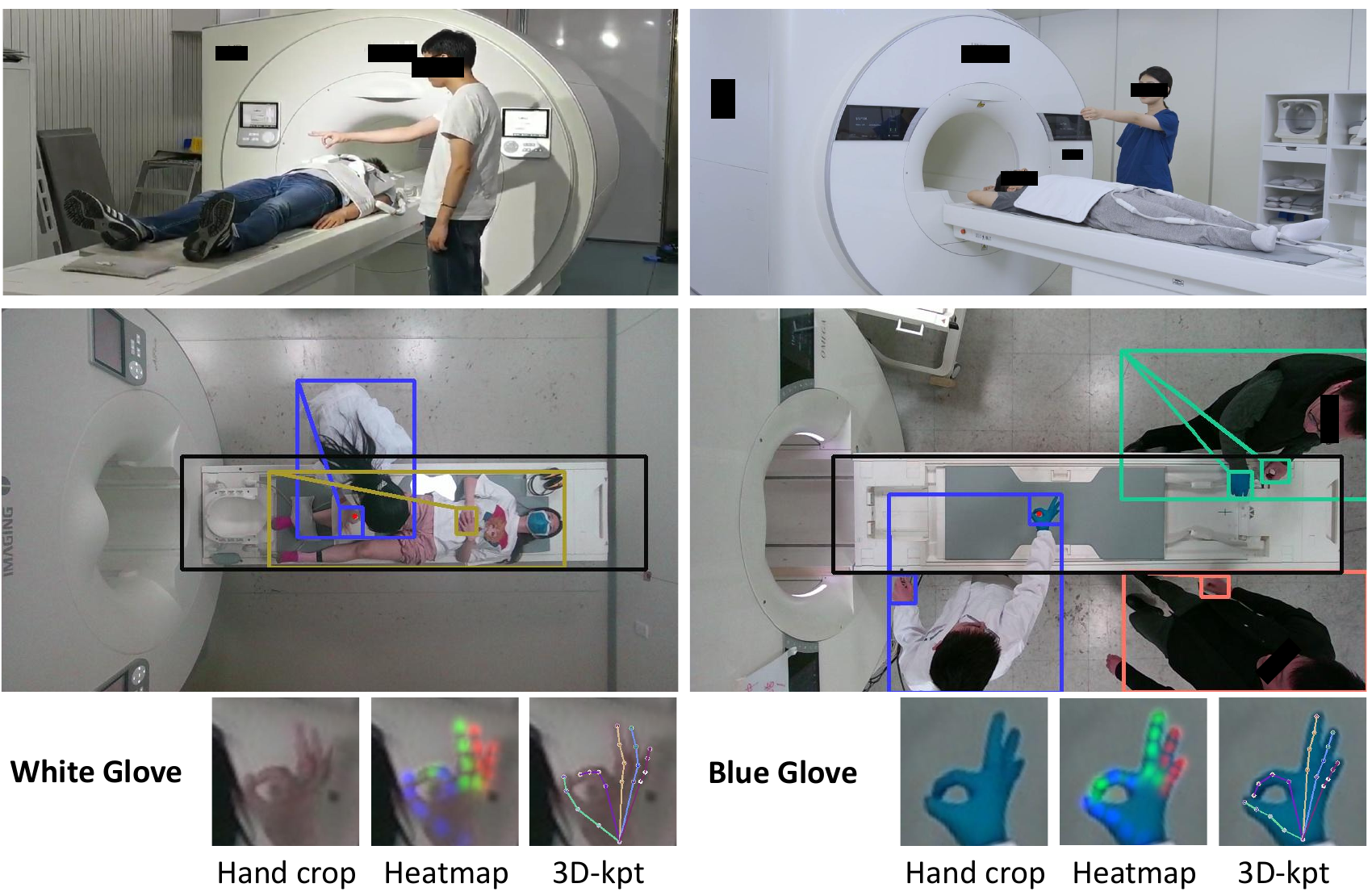}
    \caption{Gesture recognition and localization for patient positioning in scan rooms. Red dots indicate the localization areas. The same color bounding boxes represent predicted hand-body association to filter out the patient's hands. The black box is the bed.}
    \label{fig:system2}
    \vspace{-0.8em}
\end{figure*}

\subsubsection{Comparison with State-of-the-Art} 

\noindent{\textbf{Comparision with MediaPipe.}} To adapt MediaPipe hand  \cite{zhang2020mediapipe} to our task, we train a gesture classifier that takes its predicted 3D landmarks as input on our same clinical training dataset since the original MediaPipe setup cannot predict the proposed gesture. First, we infer MediaPipe hand on the cropped hand images to obtain the MediaPipe 3D landmarks (\ie, \textit{world\_coordinates}) of each hand. Then, we train an `okay/non-okay' classifier that takes the MediaPipe 3D landmarks as input, using an MLP with the same architecture as setting D in Table \ref{tab:compare}.

As shown in row F of Table \ref{tab:compare}, MediaPipe hand performs poorly in all the metrics. The hand detection in MediaPipe only has 57.6\% accuracy compared to our YOLOv7 hand detector (99.8\%). To further evaluate the hand landmark detection of MediaPipe, we use our YOLOv7 to locate the hand region and pass it to MediaPipe (row G of Table \ref{tab:compare}). The AUC improves but is still much worse than our methods which are specially designed for low-resolution and low-lighting environments. Figure \ref{fig:quality}\textcolor{red}{b} shows the qualitative comparison of landmark detection between MediaPipe and the proposed method, with the latter yielding better keypoints than MediaPipe.

We further conduct a comparison of the run-time complexity between our method and MediaPipe, as illustrated in Table \ref{tab:compare}. We can see that MediaPipe achieves the highest FPS rate. Nevertheless, our approach demonstrates comparable run-time complexity (14.4 vs. 17.4 FPS), while achieving significantly better results in terms of AUC (0.948 vs. 0.754). This indicates the superior performance and efficacy of our method. Note that, all of our testing, including the FPS measurements, is conducted directly on an NVIDIA Jetson Xavier NX edge device.

\begin{table}[tbp]
  \centering
  \resizebox{1\textwidth}{!}{
    \begin{tabular}{l|rr}
    \toprule
         Model & \multicolumn{1}{l}{mAP@0.5} & \multicolumn{1}{l}{mAP@0.5-0.95} \\
    \midrule
    HaGRID \cite{kapitanov2024hagrid} (pre-trained) & 0.008 & 0.002 \\
    HaGRID \cite{kapitanov2024hagrid} (fine-tuned) & 0.570  & 0.227 \\
    \textbf{Ours} & \textbf{0.972} & \textbf{0.739} \\
    \bottomrule
    \end{tabular}}%
    \caption{Comparison of hand detection on our gesture benchmark.}
    \vspace{-0.6em}
  \label{tab:hagrid}%
\end{table}%

\vspace{0.5em}
\noindent{\textbf{Comparison with HaGRID.}} We compare the hand detection accuracy with HaGRID \cite{kapitanov2024hagrid} in two settings: pre-trained and fine-tuned, as shown in Table \ref{tab:hagrid} and Figure \ref{fig:hagrid}. First, we use the best pret-trained model of HaGRID \cite{kapitanov2024hagrid} (\ie, RetinaNet ResNet-50) and evaluate it on our gesture benchmark. The pre-trained model fails to detect hands and both mAP@0.5 and mAP@0.5-0.95 are close to zero. This is because the model is only trained on HaGRID dataset where hands are much closer to the camera than in our dataset. Second, we fine-tune the HaGRID model in our training set. The detection accuracy improves significantly. However, our hand detection model performs the best in both two metrics. Figure \ref{fig:hagrid} shows the detection results with a confidence threshold of 0.25 and IoU threshold of 0.45. Our method shows the best detection accuracy.

\subsubsection{Evaluation on the HaGRID dataset}

To further evaluate our method, we conduct experiments on the HaGRID \cite{kapitanov2024hagrid} dataset which includes 18 gesture classes in total. First, we crop the hand images from the HaGRID dataset. Second, we train our two-layer MLP gesture classifier on the cropped training dataset using the same settings as HAGRID \cite{kapitanov2024hagrid} (\ie, SGD optimizer with a learning rate of 0.1 and a batch size of 128) for 10 epochs. As shown in Table \ref{tab:hagrid2}, our gesture classifier archives a higher F1-score than HaGRID (98.3 vs. \textbf{99.7}). At last, our whole pipeline: hand detection, 2D landmark detection, and dual-modality 3D hand pose and gesture classification, achieves higher mAP (79.1 vs. \textbf{85.0}) for hand detection and gesture recognition compared with HaGRID. 

\begin{table}[tbp]
  \centering
\captionsetup{belowskip=-2em}
  \caption{Comparison of gesture detection and classification on the HaGRID   \cite{kapitanov2024hagrid} dataset with 18 gesture classes.}
  \resizebox{1\textwidth}{!}{
    \begin{tabular}{l|ccc}
    \toprule
    Model & Parameters (M) & F1-score & mAP \\
    \midrule
    HaGRID \cite{kapitanov2024hagrid} (classifier) & 23.2  & 98.3  & - \\
    HaGRID \cite{kapitanov2024hagrid} (detection) & 38.2  & -     & 79.1 \\
    \textbf{Ours} (classifier) & 5.3   & \textbf{99.7} & - \\
    \textbf{Ours} (full) & 45.8+9.6+5.3 &    -   & \textbf{85.0} \\
    \bottomrule
    \end{tabular}}%
  \label{tab:hagrid2}%
  \vspace{-0.5em}
\end{table}%

\subsubsection{Qualitative examples} 
Figure \ref{fig:system2} illustrates examples of gesture recognition and localization for patient positioning in scan rooms. The first row of Figure \ref{fig:system2} shows two real examples of technicians using \textit{``okay''} gestures for patient positioning. The second and third rows show our model detects, recognizes, and locates white and blue gloved gesture hands. Red dots indicate the localization areas. The same color bounding boxes represent the predicted hand-body association to filter out the patient's hands. Our model can accurately associate the technician's hand with its body, recognize the gesture, and locate the ``okay'' for automated patient positioning.   

\section{Discussion and Conclusion}

In this paper, we introduce a novel workflow and analytics pipeline for automatically detecting and recognizing gestures from technicians during medical scans, to automate patient positioning. Our system uses a ceiling-mounted RGB-Depth camera, relying solely on the RGB component for 3D hand gesture recognition, and is extensively tested in a realistic clinical environment. Additionally, our 3D hand gesture recognition method is evaluated on the HaGRID public dataset, achieving state-of-the-art performance.

Future work will include clinical trials to quantify the potential benefits and compare our system's performance with traditional manual positioning methods, allowing us to validate our method's impact on efficiency, accuracy, and patient comfort.

{\small
\bibliographystyle{ieee_fullname}
\bibliography{egbib}
}

\end{document}